\documentclass[final]{cvpr}

\usepackage{times}
\usepackage{epsfig}
\usepackage{graphicx}
\usepackage{amsmath}
\usepackage{amssymb}

\usepackage[pagebackref=false,breaklinks=true,colorlinks,bookmarks=false]{hyperref}

\def\cvprPaperID{1325} % *** Enter the CVPR Paper ID here
\def\confYear{CVPR 2021}

\begin{document}

%%%%%%%%% TITLE
\title{MUST-GAN: Multi-level Statistics Transfer for Self-driven Person Image Generation}

\author{Tianxiang Ma$^{1,2}$, Bo Peng$^{2,3}$, Wei Wang$^{2}$, Jing Dong$^{2*}$\\

$^{1}$ School of Artificial Intelligence, University of Chinese Academy of Sciences\\

$^{2}$ Center for Research on Intelligent Perception and Computing, CASIA\\

$^{3}$ State Key Laboratory of Information Security, IIE, CAS \\

% $^{2}$ National Laboratory of Pattern Recognition, Chinese Academy of Sciences\\

{\tt\small tianxiang.ma@cripac.ia.ac.cn, \{bo.peng, wwang, jdong\}@nlpr.ia.ac.cn}

}

\maketitle
\pagestyle{empty}
\thispagestyle{empty}

%%%%%%%%% ABSTRACT
\begin{abstract}
 Pose-guided person image generation usually involves using paired source-target images to supervise the training, which significantly increases the data preparation effort and limits the application of the models. To deal with this problem, we propose a novel multi-level statistics transfer model, which disentangles and transfers multi-level appearance features from person images and merges them with pose features to reconstruct the source person images themselves. So that the source images can be used as supervision for self-driven person image generation. Specifically, our model extracts multi-level features from the appearance encoder and learns the optimal appearance representation through attention mechanism and attributes statistics. Then we transfer them to a pose-guided generator for re-fusion of appearance and pose. Our approach allows for flexible manipulation of person appearance and pose properties to perform pose transfer and clothes style transfer tasks. Experimental results on the DeepFashion dataset demonstrate our method's superiority compared with state-of-the-art supervised and unsupervised methods. In addition, our approach also performs well in the wild.
 
%  our model uses two encoders to extract features from person and pose image, respectively, and then utilizes a special multi-level statistics transfer network to transfer the appearance attributes from input person image into pose-guided generator, finally reconstructs the person image itself. In the process the model can spontaneously learn multi-level appearance attributes from the person images pathway, and blocking out the effects of pose attributes. Our approach can implement complex tasks such as pose transfer and clothes style transfer at the same time in a self-driven manner only. Experimental results show that the proposed method is able to achieve state-of-the-art, and exceed the models under the supervision of paired data.
\end{abstract}

\let\thefootnote\relax\footnotetext{* Corresponding author.}

%%%%%%%%% BODY TEXT
\section{Introduction}

\begin{figure}[t]
\begin{center}
\includegraphics[width=1.0\linewidth]{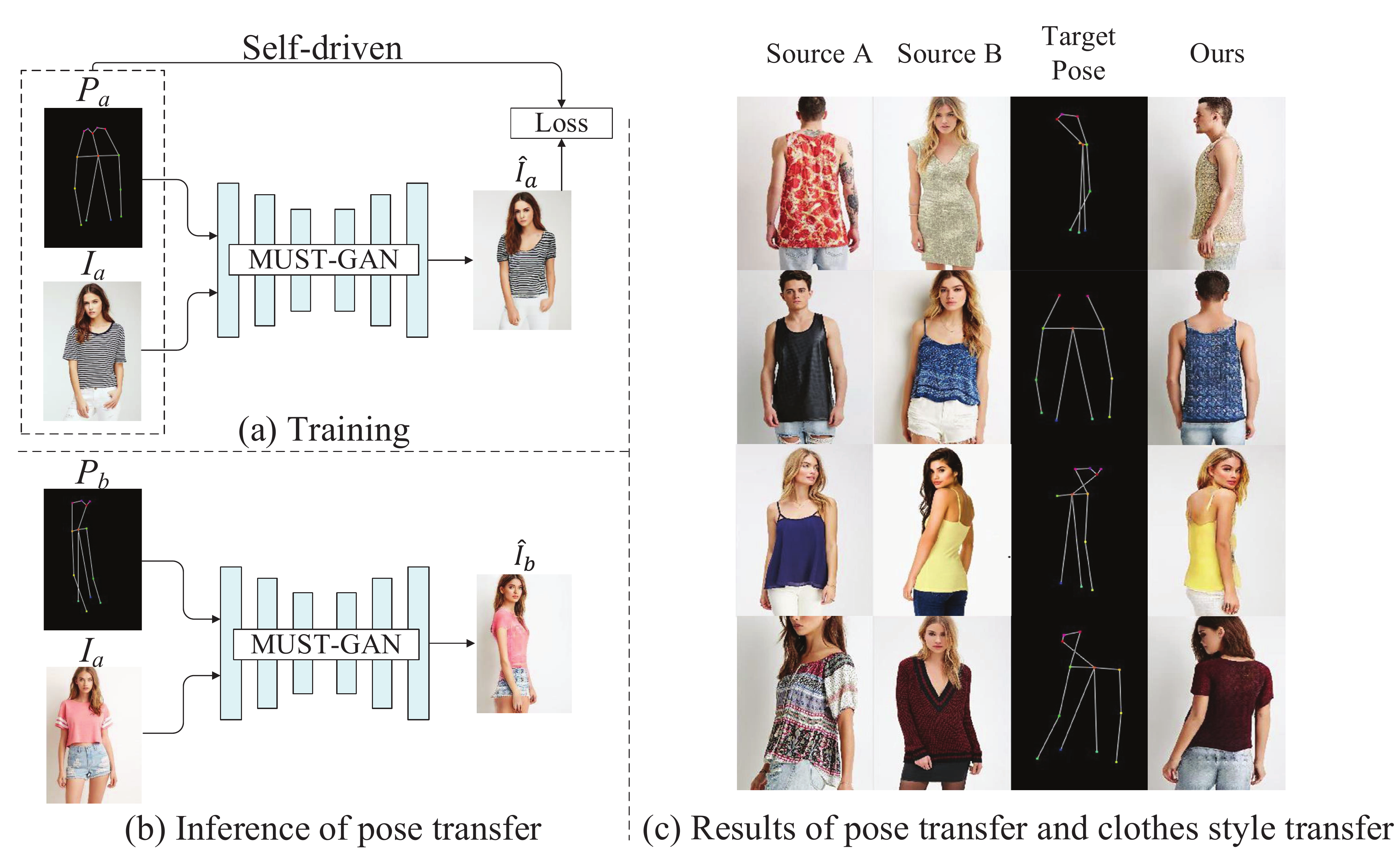}
\end{center}
   \caption{Self-driven person image generation and visualization of pose transfer with clothes style transfer. Our model can be trained in a self-driven way without paired source-target images and flexibly controls the appearance and pose attributes to achieve pose transfer and clothes style transfer in inference. The images in (c) show the generated results using this model for simultaneous pose and cloths style transfer. Source A is transferred to the target pose, and its clothes are replaced by source B's.}
\label{fig1}
\end{figure}

Person image generation has been gaining attention in recent years, which aims to generate the person image as realistic as possible, and at the same time, to transfer the source person image to a target pose. It has great potential for applications, like virtual try-on, clothing texture editing, controllable person manipulation, and so on.

Recently, many researchers have contributed to this topic, with most of the work focusing on pose-guided person image generation \cite{ma2017pose, siarohin2018deformable, pumarola2018unsupervised, zhu2019progressive, song2019unsupervised, li2019dense, men2020controllable, ren2020deep}. Many person image generation models \cite{ma2017pose, siarohin2018deformable, zhu2019progressive, li2019dense, men2020controllable, ren2020deep} use paired source-target images for supervised training. However, the paired images require a lot of time and workforce to collect, limiting the usage scenarios of these models. There are also some unsupervised person image generation methods \cite{pumarola2018unsupervised, esser2018variational, song2019unsupervised}, but the quality of their generated images is not fine.

In this paper, we propose a self-driven approach for person image generation without using any paired source-target images during training, as shown in Figure \ref{fig1}(a). Specifically, we propose a novel multi-level statistics transfer model, which can disentangle and transfer multi-level appearance features for person images. The source image can be used as supervision information for person image generation without paired source-target training data. Our method allows for flexible manipulation of pose and appearance properties to achieve pose transfer and clothes style transfer, as shown in Figure \ref{fig1}(c).

% How to disentangle person appearance features from person images and transfer them as far as possible to the pose-guided generator to reconstruct source person images is the key of our method.

Since there is no need to pair data during training for our approach, how to extract the person appearance features from input images and transfer them to a pose-guided generator for reconstruction is the key to our proposed method. It is also important to prevent the model from learning trivial solutions that directly copy all input information to generate output image. To deal with these problems, we propose a multi-level statistics transfer model, which extracts the multi-level features from the appearance encoder, and utilizes attention mechanism and attributes statistics to learn optimal representations of appearance features. Finally, the model fuses appearance features into a pose-guided generator to reconstruct the source person image.

% The appearance features are usually learned from person images, which also contain structure information of poses. How to prevent the model from learning trivial solutions that directly copy all input information to generate output image is the key problem.

% In this process, multi-level statistics extraction and transfer make more abundant appearance information to be used. The statistics of feature maps can represent its style \cite{huang2017arbitrary} and is independent of the spatial structure, which  The experimental results show its excellence in maintaining the appearance attributes of the source person image. . We extract multi-level statistics and introduce the attention mechanism to learn better feature representations. 

% which is very suitable for our task. However, the previous method only extracts the statistics at the last level of the style encoder and transfer them to the content image without trade-offs, which makes the model only exploit the high-level semantic information of the style image, and it is very inappropriate for images such as the person, where the local styles are obvious differences. 

Specifically, our method uses an appearance encoder and a pose encoder to extract features from person image and pose image, respectively. Then we introduce the MUST module to obtain multi-level statistics from the appearance encoder and use the channel attention mechanism \cite{hu2018squeeze} to learn the weights of each channel in multi-level feature maps. After that, we calculate the statistics of feature maps and apply a multi-layer fully connected network to learn the corresponding relationship when the statistics are transferred to the generator branch. In addition we propose a multi-level statistics matching network for pose-guided generator, which is composed of statistics matching residual blocks with AdaIN \cite{huang2017arbitrary} and learnable skip connection. This generator module can match the scale and channel number of multi-level statistics and generate realistic person image.

Our method is self-driven by source person images throughout the training process, and no paired source-target images are used. We compare our approach with state-of-the-art supervised and unsupervised methods. Both quantitative and qualitative comparisons prove the superiority of our method. Our model allows for flexible manipulation of person appearance and pose properties to perform pose transfer and clothes style transfer tasks in the inference, and it also shows good performance in the wild.

To summarize, the main contributions of this paper are as follows:
\begin{itemize}
  \item We propose a fully self-driven person image generation method which requires no paired source-target images for training. 
  
  \item We propose a multi-level statistics transfer model that can effectively disentangle the rich appearance features from person images and allow for flexible manipulation of person appearance and pose properties.
  
  \item Our proposed model performs well compared with the state-of-the-art methods on pose transfer and clothes style transfer and also is tested in the wild for its potential applications.
\end{itemize}

\begin{figure*}
\begin{center}
\includegraphics[width=0.90\linewidth]{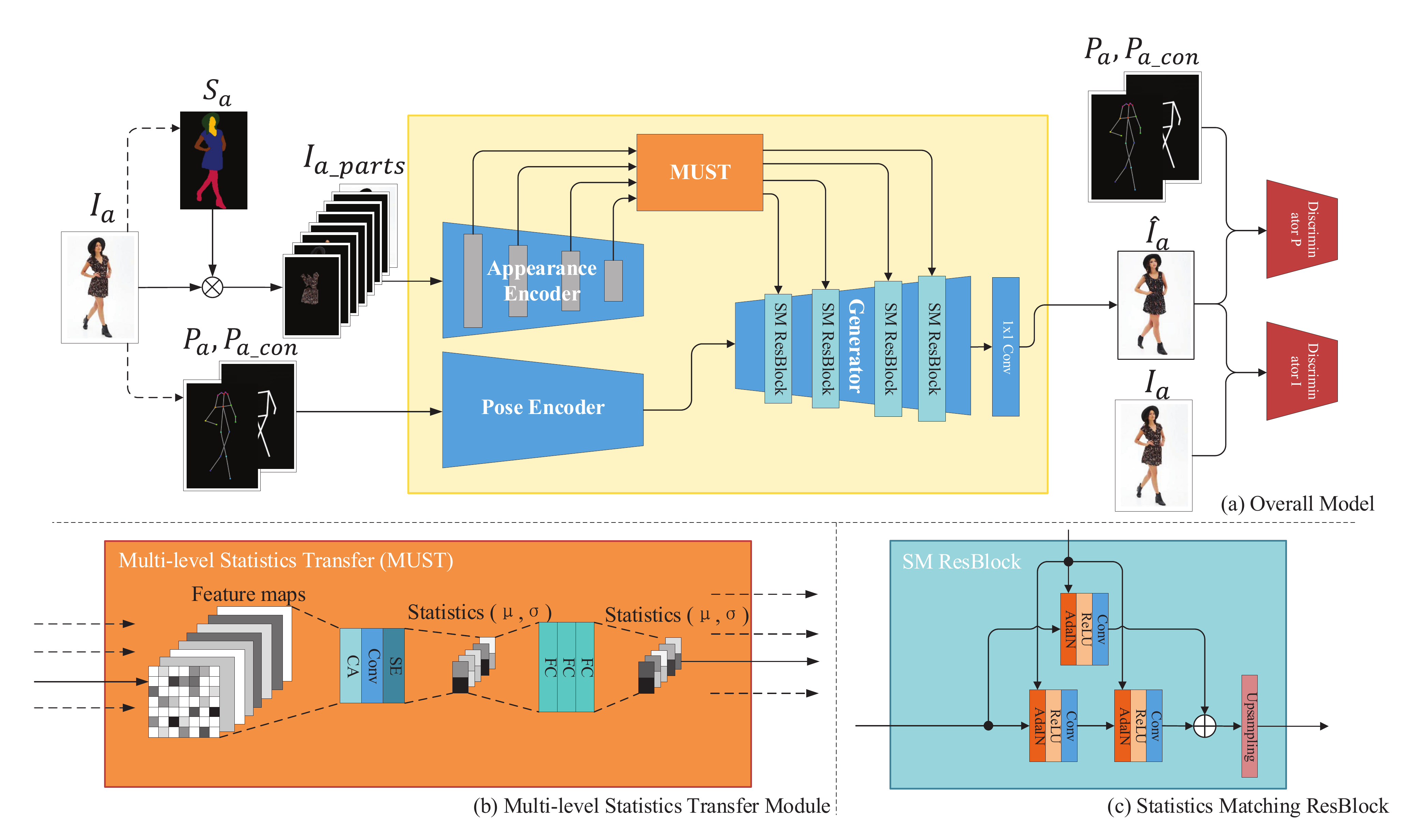}
\end{center}
   \caption{Overview of our MUST-GAN model for self-driven person image generation. Appearance encoder extracts the features of the person image parts $I_{a\_parts}$ by semantic segmentation map $S_{a}$. Pose encoder encodes the pose image $P_{a}$ and pose connection map $P_{a\_con}$ and guides the Generator to synthesize the source posture. The MUST module disentangles and transfers multi-level appearance features, and the Generator fuses the multi-level appearance features and pose codes for reconstruction of the source person image $ I_{a}$.}
\label{fig2}
\end{figure*}

\section{Related Work}
\subsection{Image Generation}
Thanks to the emergence and development of generative adversarial network \cite{goodfellow2014generative}, image generation models have been developing at a high rate in recent years. Some methods \cite{radford2015unsupervised, karras2018progressive, karras2019style, karras2020analyzing} use random noise as the input of the network, and the others \cite{mirza2014conditional, isola2017image, zhu2017unpaired, wang2018high} use conditional information as the input. Currently, the generative models with conditional inputs can generate more specific and controllable images. Pix2pix \cite{isola2017image} can generate a street view or object of a particular shape based on different semantic or edge images. CycleGAN \cite{zhu2017unpaired} implements image translation in an unsupervised manner through cyclic reconstruction. Many methods are devoted to improving the resolution and quality of the generated images. StyleGAN \cite{karras2019style} utilizes progressive generation to enhance image resolution and uses AdaIN \cite{huang2017arbitrary} embedded style code to control the style of the generated images. Pix2pixHD \cite{wang2018high} improves the quality of the generated images using a two-stage generation from coarse to fine, and further improves the model using a multi-scale discriminator. What's more, some work focuses on improving the generative model with stacked architecture \cite{zhang2017stackgan}, attention mechanism \cite{zhang2019self}, and latent representation \cite{liu2017unsupervised, huang2018multimodal}. Besides, some work starts to explore the few-shot learning image translation \cite{liu2019few, saito2020coco}.

% which utilizes a two-stage generation approach to synthesis person images from coarse to fine and uses a U-Net-like network as the structure of the generator.

\subsection{Pose-guided Person Image Generation}
Pose-guided person image generation is an important sub-area of image generation that has been continuously developed in recent years. Ma et al. \cite{ma2017pose} proposed the first method for pose-guided person image generation. Siarohin et al. \cite{siarohin2018deformable} utilized affine transformation to model the process of pose transfer. Esser et al. \cite{esser2018variational} employed a variational autoencoder \cite{kingma2013auto} and combined with conditional U-Net \cite{ronneberger2015u} to model the shape and appearance of person images. Zhu et al. \cite{zhu2019progressive} introduced a progressive pose attention transfer network to build the model structure in the form of residual blocks. Liu et al. \cite{liu2019liquid} warped the input images at the feature level with an additional 3D human model. Li et al. \cite{li2019dense} added an optical flow regression module to the usual U-Net-like person image generation model to guide the pose transformation. Han et al. \cite{han2019clothflow} used a flow-based method to transform the source images at the pixel level. Men et al. \cite{men2020controllable} used person parsing to divide the person image into semantic parts and fed them into a shared parameter encoder to get the style code. Besides, Ren et al. \cite{ren2020deep} considered both global optical flow and local spatial transformations to generate higher quality images. Tang et al. \cite{tang2020xinggan} applied a co-attention mechanism to shape and appearance branches, respectively. All of these methods supervise the train of models and require paired source-target images of person and pose. Some methods began to consider unsupervised person image generation. Pumarola et al. \cite{pumarola2018unsupervised} first proposed unsupervised person image generation, which uses the CycleGAN to achieve unsupervised pose transfer, but the images generated in this way are not detailed enough. Song et al. \cite{song2019unsupervised} also used CycleGAN and broke down the person pose transfer into semantic parsing transformation and appearance generation, to reduce the difficulty of the direct pose transfer. However, both methods require a target pose to guide the transformation. Esser et al. \cite{esser2019unsupervised} introduced disentangle representations of appearance and pose, which doesn't require pose annotations but needs pairs of images depicting the same object appearance. In this paper, our method can not use any target information, only rely on the input images to supervise the model training, to achieve self-driven person image generation.

% \subsection{Disentangled Representations}
% Learning disentangled Representations of images has always been a research focus, which aims at modeling the factors of data variations. Many methods 

% \subsection{Style Transfer}
% Pose-guided person image generation can be seen as a special case of image style transfer, where person images and pose images can be analogous to style images and content images. Gatys et al. \cite{gatys2016image} first achieved attractive style transfer results by matching feature statistics in the CNN. Later, feed-forward style transfer  approaches \cite{johnson2016perceptual, ulyanov2016texture, ulyanov2017improved} are proposed to greatly accelerate image style transfer and further improve transferred quality. However, these methods are limited into a fixed style. Some methods \cite{li2017diversified, dumoulin2016learned} attempted to make the style transfer model suitable for various style images. On the other hand, there is work beginning to explore general style transfer. Huang and Belongie \cite{huang2017arbitrary} proposed an adaptive instance normalization(AdaIN) layer that can better align the mean and variance of feature maps between style and content images. Image feature map statistics such as mean and variance have the advantages of texture representation and independence with spatial structure. Therefore, we will exploit these characteristics to build person image generation model in this paper.

\section{Approach}
In this paper, we want our method to effectively disentangle the person's appearance and pose properties so that the input person images and pose images are independent in the model, and the model only needs to learn the feature representation and feature fusion of the person and pose images. Thus the input images can be arbitrary, i.e., we can use only the source person image and the corresponding pose to realize self-driven person image generation. To achieve this goal, we propose a multi-level statistics transfer model, as shown in Figure \ref{fig2}. It contains four essential parts. Two pathway encoders for person appearance and pose, respectively, a multi-level statistics transfer network (MUST) and a multi-level statistics matching generator network.

Formally, we define $I_{a}$, $I_{b}$ represent person images, and $P_{a}$, $P_{b}$ represent the corresponding postures, where $a$ and $b$ represent source and target poses of the same person wearing the same clothes. Models with paired data supervision can be written as $\hat{I_{b}} = G(I_{a},P_{b})$, where $G$ is the generator, $\hat{I_{b}}$ is the generated person image in the target pose $b$, and the training set is $\{I_{a},P_{b},I_{b}\}$. For the previously unsupervised person image generation models \cite{pumarola2018unsupervised, song2019unsupervised}, the formula is $\hat{I_{a}} = G(I_{a}, P_{a}, P_{b})$, and the training set is $\{I_{a}, P_{a}, P_{b}\}$, where the model needs the target pose $P_{b}$ as input. However, our proposed model realizes the self-driven person image generation without any target images' information, formalized as $\hat{I_{a}} = G(I_{a}, P_{a})$, and the training set is $\{I_{a}, P_{a}\}$, where all supervision information is derived from the source images.

\subsection{Pose Encoder}
In the pose pathway, we utilize the trained person pose estimation method \cite{cao2017realtime} to get person pose joints estimate and construct the pose joints as 18-channel heat maps $P_{a}$. There are connections between body joints, such as arms and legs, while person postures represented by heat maps lack such connections. Therefore, we introduce a pose connection map $P_{a\_con}$, including the trunk and limbs, which is spliced with the joint heat maps along the channel dimension, to help the model generate a more accurate structure of the person. For the pose encoder network, we use a down-sampling convolutional neural network with Instance Normalization to encode the joint heat maps and pose connection maps into a high-dimensional space to guide the generator network.

\subsection{Appearance Encoder}
For the person images pathway, in order to the preliminary disentanglement of the person image, we use the person parsing \cite{gong2017look} to obtain the semantic segmentation maps $S_{a}$ of the input person image. Then we merge the semantic maps into eight classes as in  \cite{men2020controllable} and element-wise multiply them with the person image to obtain $I_{a\_parts}$. It enables complex person appearance to be segmented into several parts to facilitate feature extraction and transfer of the network later. It is also useful for the task of clothes style transfer. The appearance encoder's purpose is to extract rich and robust features at different levels to serve the later MUST module. Therefore, we use the VGG \cite{simonyan2014very} model trained in the COCO dataset \cite{lin2014microsoft} as the appearance encoder. The advantage of this is that robust and rich image features can be obtained at the early stage of model training. Because the VGG model is trained on COCO, its generalization is better, and it facilitates the application of our model in the wild.

% We construct the extracted features as feature pyramids,
% \begin{equation}
% {f}_{i}=R\left( E_{I}\left(I_{a}\right), Conv_{i} \right), i \in {1,2,...,N}
% \end{equation}
% where $E_{I}$ is the feature extractor, $Conv_{i}$ represents the output of the i-th VGG convolutional layer, and $R$ is a feature selection and reshape operation used to select a specific VGG output layer and make the number of extracted feature channels match the input of the MUST module. $f_{i}\in R^{C_{i},H_{i},W_{i}}$ is the extracted i-th level features. 

% The purpose of constructing a feature pyramid is to obtain person image features at different scales and semantic levels(e.g. low level texture and color, high level style), including both appearance and pose features, and the function of the MUST module is to extract and transfer the appearance features while blocking out the pose features.

\subsection{Multi-level Statistics Transfer}
Our model is trained without using paired source-target images, as we find that when effectively disentangled and transfer person appearance and pose features, the source image itself can provide supervisory information to guide person image generation. We call this self-driven person image generation. 

To achieve effective decoupling and transfer of appearance attributes, we propose a multi-level statistics transfer network, as shown in Figure \ref{fig2}(b). We select features extracted from the appearance encoder at four levels from shallow to deep. And first, we utilize the channel attention layer to learn adaptive weights for the features at each level and then reduce the number of channels of the feature maps to a size suitable for the generator network through the convolution layer. After that, we extract the statistics(mean and variance) of each feature map. Because the statistics are the global information, the feature map's structural information is masked, while the statistics can represent the style information. This method enables the disentanglement of appearance and pose in the person image. The use of multi-level features allows the extracted person appearance attributes to have low-level color and texture information and high-level style information.

% In the style transfer method \cite{huang2017arbitrary}, the authors transfer the statistics(mean and variance) of the feature maps from the style image to the content image, where the feature maps are derived from the high-level outputs of the feature extractor network. Since the high-level features contain the image global structure and style information, and the extracted statistics are representative of the style attributes, this method can shift the image style while preserving the image content. In pose-guided person image generation, similarly, the pose image is the content image and the person image is the style image. However, in the pose transfer, the generated person image is not only the style of the high-level features, but also the color and texture of the lower-level features to be identical to the source image. For the consideration that the features of each level should be transferred, we propose a multi-level statistics transfer module, which can transfer the local colors and textures as well as the overall style of the person image, and ensure that these attributes are consistent with that of the source image when generating person images of other poses. 

\begin{figure}[t]
\begin{center}
\includegraphics[width=0.74\linewidth]{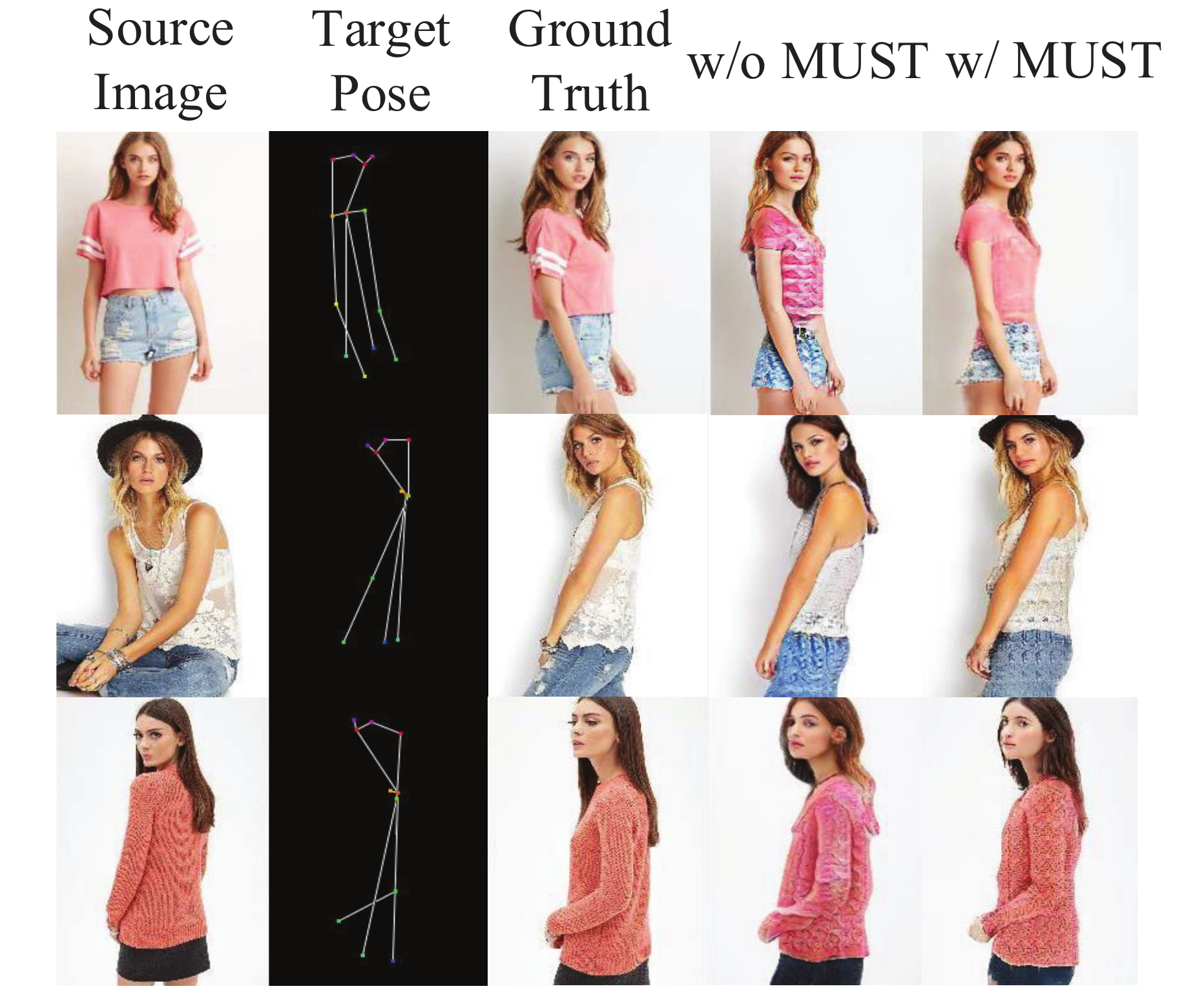}
\end{center}
   \caption{The effect of the MUST module in our model.}
\label{fig3}
\end{figure}

The effect of this module is shown in Figure \ref{fig3}. From the results, we can see that using the MUST allows the model to acquire and transfer more accurate appearance attributes. The images generated without the MUST are very different in color and texture from the source image. More specific quantitative comparisons are discussed in the ablation study experiment. The MUST can be simply expressed as
\begin{equation}
{s}_{i}=SE\left( Conv_{i}\left( CA\left(f_{i}\right)\right)\right), i \in {1,2,3,4},
\end{equation}
where $f_{i}$ is the feature maps, $CA$ is the channel attention layer, and $SE$ is the statistics extraction operation. After this, we utilize a multi-layer fully connected network to transform the extracted attributes statistics so that the network learns the mapping of the statistics in the generator. So the complete MUST network can be represented as
\begin{equation}
{s}_{i}=Trans\left(SE\left( Conv_{i}\left( CA\left(f_{i}\right)\right)\right)\right), i \in {1,2,3,4},
\end{equation}
where the $Trans$ represents feature transformation of a multi-layer fully connected network.

% including mean and variance, MLP is the level-wise multilayer perceptron network used for adaptive transforming statistics parameters, and $p_{i}$ is the output statistics, which can better represent the appearance of the person image and better adapt to the later generator network. In addition to represent and transfer the appearance of person images very well, MUST more importantly solves the problem of person image generation under completely unpaired images, i.e., attributes disentanglement. Because the attributes statistics extracted by MUST are the global information of each feature level, including the color and texture at the low level and the style at the high level. What's more, the statistics are global and do not contain any spatial structure information, they can naturally block the influence of the pose attribute in feature maps, thus preventing the model from learning a degenerate solution. 

\subsection{Multi-level Statistics Matching Generator}
Since we have extracted attributes statistics at multiple levels by MUST, we need to map the statistics to the pose-guided generator and correctly match the features' size and channel at each level. Therefore, we propose a multi-level statistics matching generator network, which is composed of four statistics matching residual blocks, as shown in Figure \ref{fig2}(c). First of all, the attributes statistics parameters obtained from MUST are applied to the generator through the AdaIN \cite{huang2017arbitrary}, which normalizes the network features and adjusts the distribution of feature maps according to the input statistics parameters. Secondly, we use a multi-level residual blocks network as the generator's backbone. And we utilize a learnable skip connection to implement the residual structure even if the numbers of input and output channels are different. The bilinear upsampling layer is used to increase the size of the feature map gradually. Finally, the person image is reconstructed by 1$\times$1 convolutional layer, which integrates each channel's features.

\subsection{Discriminator}
There are two commonly used discriminators for pose-guided person image generation methods \cite{pumarola2018unsupervised, zhu2019progressive, men2020controllable}, $D_{I}$ for person images and $D_{P}$ for pose images. Similarly, we utilize both discriminators. But for $D_{P}$, we add a pose connection map input $P_{a\_con}$, which provides the interrelationship of the pose joints to produce a more accurate posture of the person image. We splice it with the original 18-channel joint heat maps $P_{a}$ along channel dimension. Both discriminators use residual blocks and downsampling convolutional layers similar to the method \cite{zhu2019progressive}.

\subsection{Loss Functions}
The purpose of our model's overall loss functions is to make the generated person image conform to the input person image in appearance and match the input pose image in posture. The specifics are as follows, where $I_{a}$ and $P_{a}$ in all formulas represent the input person image and pose image, $P_{a\_con}$ and $P_{a}$ are merged together as $P_{a}$, $G(I_{a}, P_{a})$ represents the generated person image.

\noindent \textbf{Adversarial loss.} We use the lsgan \cite{mao2017least} loss as the adversarial loss. Both discriminator $D_{I}$ and $D_{P}$ are used to help the generator $G$ to synthesize realistic person image in a particular pose and keep the input person image's appearance. Adversarial loss is defined as:
\begin{equation}
\begin{aligned}
\mathcal{L}_{\text{adv}}=&\mathbb{E}_{I_{a}, P_{a}}\left[\log \left(D_{I}\left(I_{a}\right) \cdot D_{P}\left(P_{a}, I_{a}\right)\right)\right]+\\
& \mathbb{E}_{I_{a}, P_{t}}\left[\log \left(\left(1-D_{I}\left(G\left(I_{a}, P_{a}\right)\right)\right)\right.\right.\\
&\left.\left.\cdot\left(1-D_{P}\left(P_{a}, G\left(I_{a}, P_{a}\right)\right)\right)\right)\right].
\end{aligned}
\end{equation}

\noindent \textbf{Reconstruction loss.}
The reconstruction loss aims to make the generated person image match the input person image at the pixel level. Reconstruction loss is calculated using the L1 loss function with the following formula:
\begin{equation}
\mathcal{L}_{\text{rec}}=\left\|G\left(I_{a}, P_{a}\right)-I_{a}\right\|_{1}.
\end{equation}

\noindent \textbf{Perceptual loss.} Unlike reconstruction loss, the perceptual loss constrains the generated image at a higher feature level. This loss function comes from the style transfer method \cite{johnson2016perceptual}. Similar to it, we use the trained VGG19 network to extract the features of the generated and input person images, respectively, and compute L1 loss for the features at the specified level $l$ with the following equation,

% This loss function comes from the style transfer method \cite{johnson2016perceptual}, which uses a trained VGG network $\phi$ to extract and contrast the generated and input image's features to help better image generation.

\begin{equation}
\mathcal{L}_{\text{perc}} =\left\|\mathcal{\phi}^{l}\left(G\left(I_{a}, P_{a}\right)\right)-\mathcal{\phi}^{l}\left(I_{a}\right) \right\|_{1}.
\end{equation}

\noindent \textbf{Style loss.}
To further improve the similarity between the generated person image and the input image in terms of appearance attributes such as texture and color, we use a style loss introduced by method \cite{johnson2016perceptual}. This loss function calculates the statistic error between activation maps of the input image and the generated image with the Gram matrix. The specific formula is as follows,
\begin{equation}
\mathcal{L}_{\text {style}}=\sum_{l}\left\|\mathbb{G}\left(\mathcal{\phi}^{l}\left(G\left(I_{a}, P_{a}\right)\right)\right)-\mathbb{G}\left(\mathcal{\phi}^{l}\left(I_{a}\right)\right)\right\|_{1},
\end{equation}
where $\phi^{l}$ is the $l$th activation layer of the trained VGG network, and $\mathbb{G}$ is the Gram matrix.

The overall loss function is,
\begin{equation}
\begin{aligned}
\mathcal{L}_{\text{overall}}=&\lambda_{\text {adv}}\mathcal{L}_{\text{adv}}+\lambda_{\text {rec}} \mathcal{L}_{\text {rec}}+\lambda_{\text {perc}} \mathcal{L}_{\text {perc}}\\
&+\lambda_{\text {style}} \mathcal{L}_{style}
\end{aligned}
\end{equation}
where $\lambda_{\text {adv}}$, $\lambda_{\text {rec}}$, $\lambda_{\text {perc}}$, $\lambda_{\text {style}}$ are the weights of the corresponding loss functions.

\begin{figure*}[t]
\begin{center}
\includegraphics[width=1.0\linewidth]{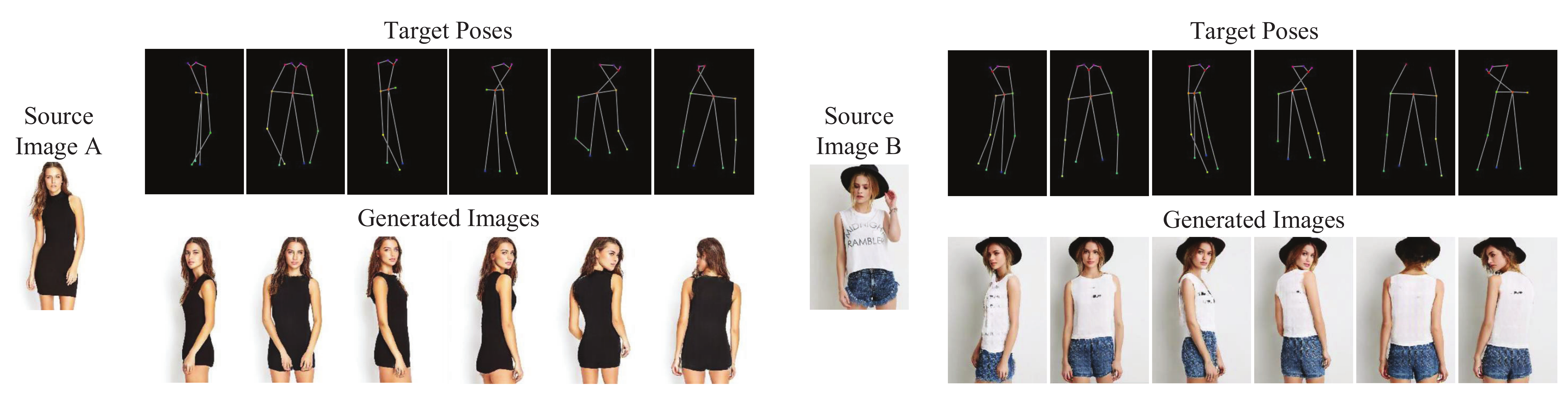}
\end{center}
   \caption{The results of our method in the pose transfer task.}
\label{fig7}
\end{figure*}

\section{Experiments}
\subsection{Implementation Details}
In this section, we specify the implementation details of our method. 

\noindent \textbf{Dataset.} We validate our method on the DeepFashion: In-shop Clothes Retrieval Benchmark \cite{liu2016deepfashion}, which contains a large number of model's person images in different poses and different appearances and is widely used in person image generation methods. This dataset includes 52,712 person images with a resolution of 256$\times$256. Since our model can be trained without paired source-target images, the cross pose pairs provided in the dataset are not required. Therefore, we randomly select 37,344 images from the entire dataset as training data. To validate applications such as pose transfer and clothes style transfer during the inference phase, we use the test data pairs of 8570 images constructed in pose transfer model \cite{zhu2019progressive}. And there is no overlap between the training set and this test set.

\noindent \textbf{Metrics.} Many previous methods of person image generation have used Structure Similarity(SSIM) \cite{wang2004image} and Inception Score(IS) \cite{salimans2016improved} as quantitative evaluation metrics for generated images, which were first introduced to the person image generation task by Ma et al. \cite{ma2017pose}. They are used to evaluate the similarity of the generated person images to Ground Truth and the realism and diversity of the generated images. To further verify the effect of the model in this paper, we also use Fréchet Inception Distance(FID) \cite{heusel2017gans} and Learned Perceptual Image Patch Similarity(LPIPS) \cite{zhang2018unreasonable} metrics to assess the realism and consistency of generated images, which have recently been applied to the evaluation of person image generation models  \cite{ren2020deep}.

% It measures the difference between the generated images and the real images by calculating the Wasserstein-2 distance between the two at the feature level.

\noindent \textbf{Network architecture.} The model proposed in this paper uses end-to-end training with an Auto-Encoder-like structure. Specifically, the pose encoder uses a downsampling convolutional neural network for encoding pose information. The appearance encoder is a trained VGG network on the COCO dataset. The multi-level statistics matching generator utilizes a residual blocks network, bilinear upsampling, and 1$\times$1 convolutional layer to generate a realistic person image. The multi-level statistics transfer network uses the structure, as shown in Figure \ref{fig2}(b).

\noindent \textbf{Training details.} Our method is implemented in PyTorch using an NVIDIA TITAN RTX GPU with 24GB memory. Pose estimation uses OpenPose \cite{cao2017realtime}, and person parsing utilizes the method \cite{gong2017look}. We group the person semantic segmentation maps extracted by  \cite{gong2017look} into eight parts, i.e., upper clothes, pants, skirt, hair, face, arm, leg, and background same with  \cite{men2020controllable}. We adopt Adam \cite{kingma2014adam} optimizer with the learning rate as $10^{-4}$ for Generator and $4\times10^{-4}$ for Discriminator (TTUR \cite{heusel2017gans}). The weights for the overall loss function are set to $\lambda_{\text {adv}}=5$, $\lambda_{\text {rec}}=1$, $\lambda_{\text {perc}}=1$, $\lambda_{\text {style}}=150$ for all experiments.

\subsection{Pose Transfer}
One of the most important applications of person image generation is the pose transfer. Given the source person image and the target pose, it is required to generate the source person image in the target pose. The results of our model in the pose transfer task are shown in Figure \ref{fig7}. From the results, we can see that our model enables pose transfer well and maintains source images' appearance information. In addition, we compare our method with a number of state-of-the-art methods, including methods \cite{siarohin2018deformable, karras2018progressive, li2019dense, men2020controllable, ren2020deep} with paired source-target images supervision, and unsupervised method \cite{pumarola2018unsupervised, ma2018disentangled, esser2018variational, song2019unsupervised}.

\begin{figure}
\begin{center}
\includegraphics[width=1.0\linewidth]{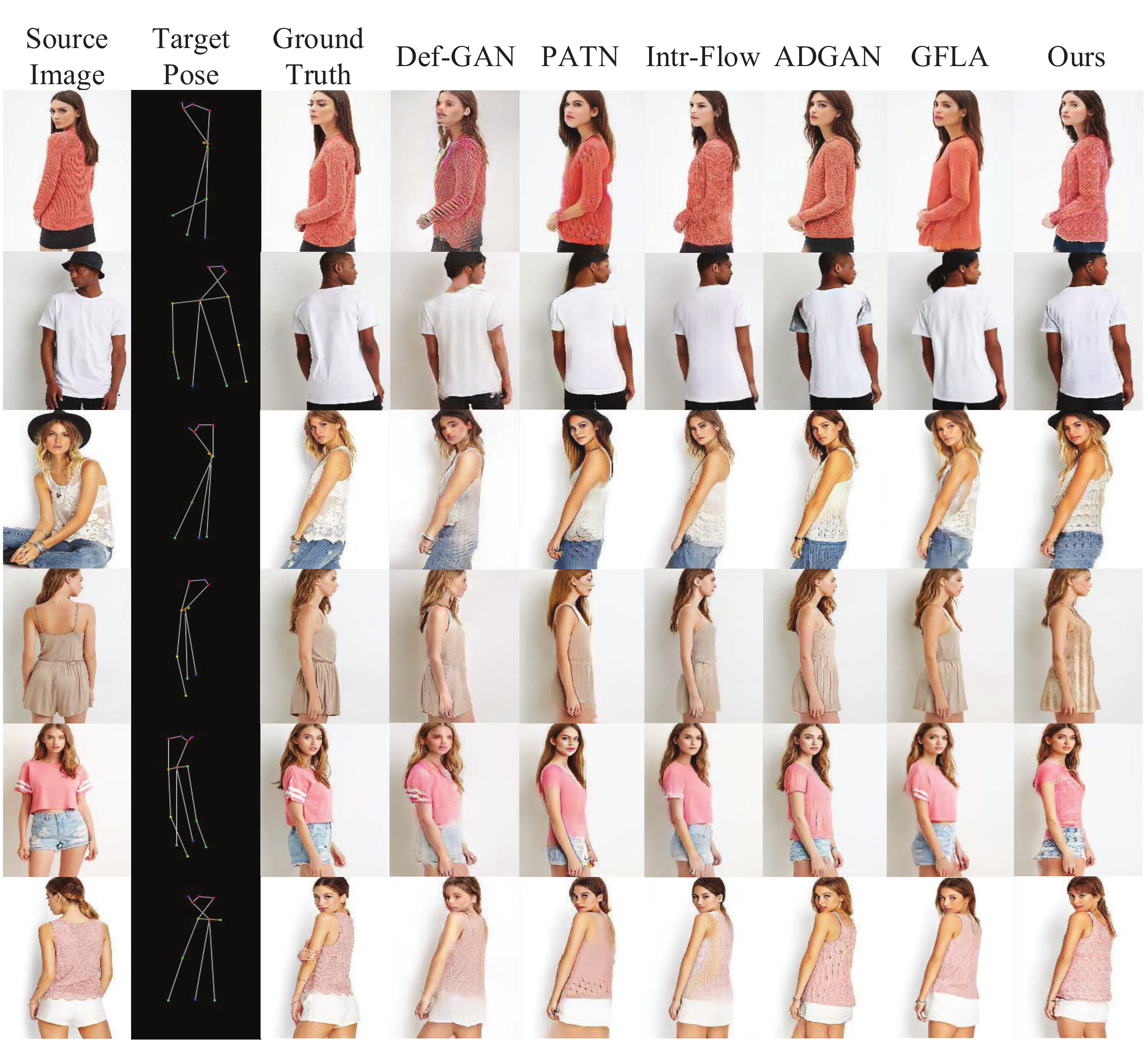}
\end{center}
   \caption{The qualitative comparison with state-of-the-art methods.}
\label{fig4}
\end{figure}

\begin{figure*}
\begin{center}
\includegraphics[width=1.0\linewidth]{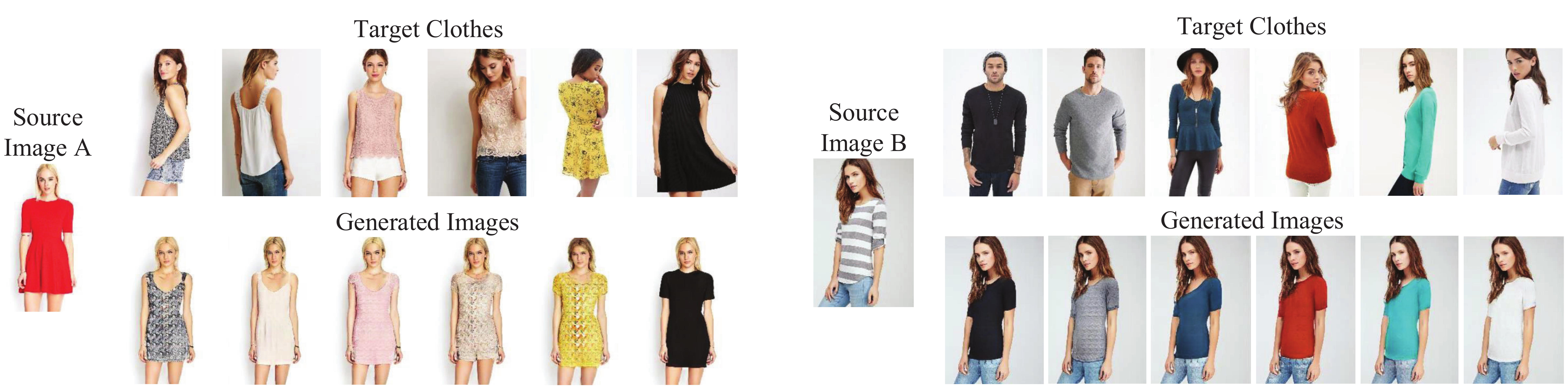}
\end{center}
   \caption{The results of our method in the clothes style transfer task.}
\label{fig8}
\end{figure*}

\subsubsection{Qualitative Comparison}
The results of the qualitative comparison are shown in Figure \ref{fig4}. We compare the generated images of our method with several state-of-the-art approaches, including Def-GAN \cite{siarohin2018deformable}, PATN \cite{karras2018progressive}, Intr-Flow \cite{li2019dense}, ADGAN \cite{men2020controllable} and GFLA \cite{ren2020deep}. All results are obtained using source code and the trained model released by the authors. We can see from the figure that the person image generated by our method is comparable to that of SOTA methods. The preservation of the source person image's appearance properties is also perfect, e.g., in the third line of the figure, our approach is able to maintain the hat feature in the source person image, while other methods do not. It is also worth noting that all the methods compared here require using complete paired source-target images to train the models. In contrast, our method's training process is more demanding because our model is self-driven, relying only on information from the source images to supervise the model. No source-target pairs of person and pose images are used during the entire training process.

\begin{table}
\begin{center}
\setlength{\tabcolsep}{2mm}{
\begin{tabular}{c|c|c|c|c}
\hline 
\text {Model} & \text {IS$\uparrow$} & \text {SSIM$\uparrow$} & \text {FID$\downarrow$} & \text {LPIPS$\downarrow$}\\
\hline 
\hline
\text { UPIS \cite{pumarola2018unsupervised} } & 2.971 & 0.747 & - & -\\
\text { BodyROI7 \cite{ma2018disentangled} } & 3.228 & 0.614 & - & -\\
\text { VU-Net \cite{esser2018variational} } & 3.087 & \textbf{0.786} & 23.583 & 0.2637\\
\text { E2E \cite{song2019unsupervised} } & 3.441 & 0.736 & 19.248 & 0.2546\\

% \hline

\text { Ours } & \textbf{3.692} & 0.742 & \textbf{15.902} & \textbf{0.2412}\\
\hline
\end{tabular}}
\end{center}
   \caption{Quantitative comparison with unsupervised state-of-the-art
methods on DeepFashion.}
\label{tab1}
\end{table}

\subsubsection{Quantitative Comparison}
First, we quantitatively compare our method with the unsupervised person image generation methods, as shown in Table\ref{tab1}. The results show that our method achieves SOTA on IS, FID, and LPIPS metrics. UPIS cannot calculate the FID and LPIPS metrics because the source code is not available. BodyROI7 uses a different test protocol from other methods and therefore cannot calculate the valid metrics of FID and LPIPS. However, comparing the other two metrics still shows the superiority of our method. What's more, it is worth noting that UPIS and E2E require the target pose as input, while our approach requires no target information and is entirely self-driven training.

To further validate our method's performance, we also compare with the supervised methods, as shown in Table \ref{tab2}. Similarly, our method performs best on the IS metrics. For SSIM and LPIPS scores, all methods are relatively close, and our approach is also at a high level for FID. Although our approach does not score as high as GFLA, which explicitly models the source-to-target transformation process, our method even exceeds most supervised models. And it is important that our method is trained in a self-driven manner without source-target pairs supervision.

\subsection{Ablation Study}
In this section, we perform an ablation experiment to validate the effect of our proposed MUST module, channel attention mechanism and pose connection map(PCM) on the overall model. For the MUST module's ablation, we remove the MUST module from the model, use the original encoder similar to that in  \cite{huang2017arbitrary}, and directly transfer the attributes statistics from the last level of the appearance encoder to the generator network. For the ablation of CA and PCM, we directly remove them from the model. The results of the experiment are shown in Tabel \ref{tab3}. From the table, we can see that the MUST module improves the model under all metrics very much. The channel attention mechanism in the MUST also plays an important role, which enhances the model under SSIM, FID and LPIPS metrics. The PCM has also improved our model on these metrics. The ablation experiments demonstrate the ability of MUST to extract and transfer appearance attributes. And the introduction of channel attention mechanism and pose connection map further enhance the realism and consistency of the images generated by our model under the evaluation metrics in this experiment.

\begin{table}
\begin{center}
\setlength{\tabcolsep}{2mm}{
\begin{tabular}{c|c|c|c|c}
\hline \text {Model} & \text {IS$\uparrow$} & \text {SSIM $\uparrow$} & \text {FID$\downarrow$} & \text {LPIPS$\downarrow$}\\
\hline 
\hline
\text { Def-GAN \cite{siarohin2018deformable} } & 3.141 & 0.741 & 18.197 & 0.2330\\
\text { PATN \cite{karras2018progressive} } & 3.213 & 0.770 & 24.071 & 0.2533\\
\text { Intr-Flow \cite{li2019dense} } & 3.251 & \textbf{0.794} & 16.757 & \textbf{0.2131}\\
\text { ADGAN \cite{men2020controllable} } & 3.329 & 0.771 & 18.395 & 0.2383\\
\text { GFLA \cite{ren2020deep} } & 3.635 & 0.713 & \textbf{14.061} & 0.2341\\

% \hline

\text { Ours } & \textbf{3.692} & 0.742 & 15.902 & 0.2412 \\
% \text { Ours (sup.) } & 3.257 & 0.784 & 15.906 & 0.2171 \\

\hline
\end{tabular}}
\end{center}
   \caption{Quantitative comparison with supervised state-of-the-art methods on DeepFashion.}
\label{tab2}
\end{table}

\begin{figure*}
\begin{center}
\includegraphics[width=0.86\linewidth]{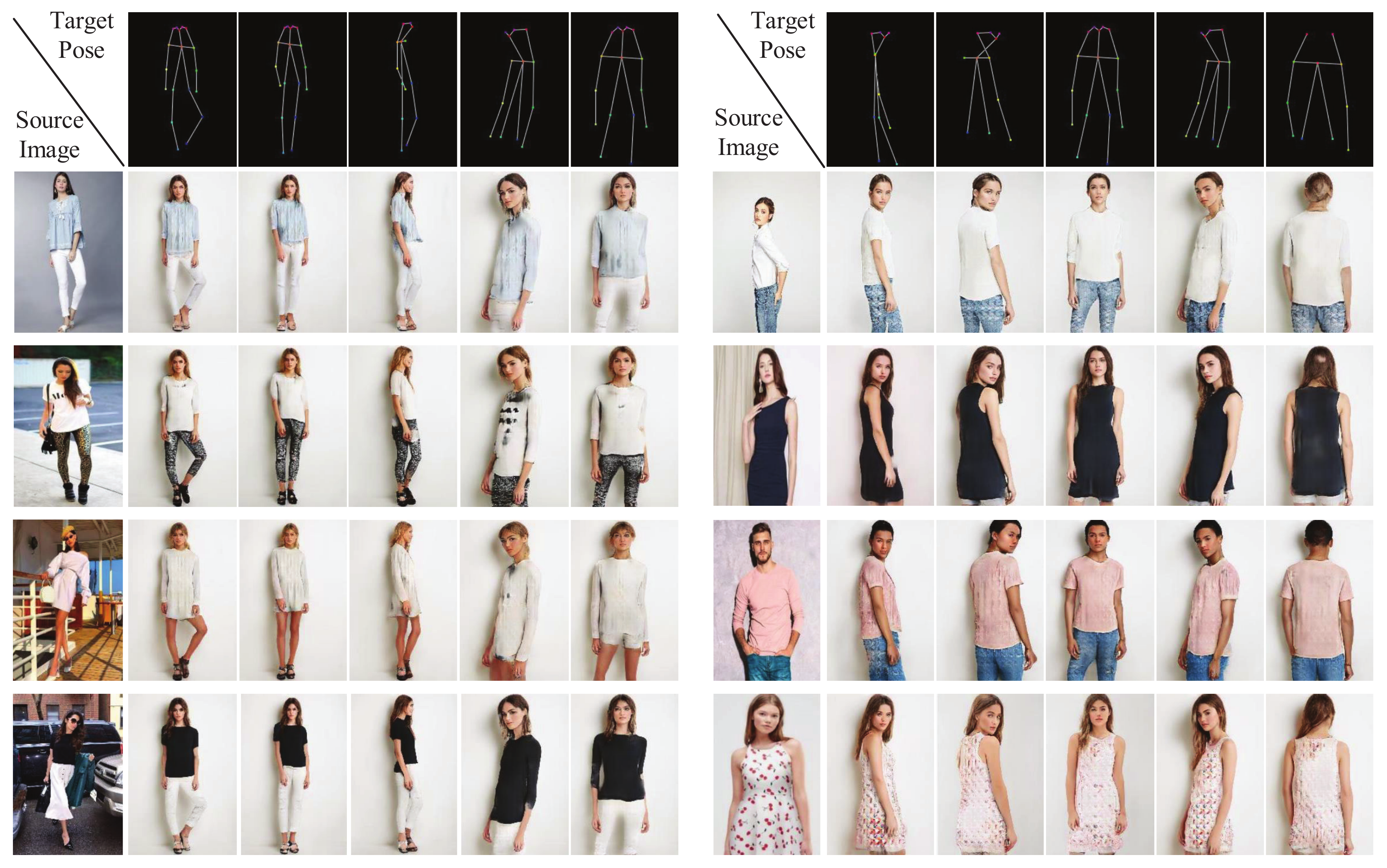}
\end{center}
   \caption{The test results of our method in the wild.}
\label{fig5}
\end{figure*}

\subsection{Clothes Style Transfer}
Our proposed method can manipulate the appearance and pose attributes of the person images separately. So the model can achieve clothes style transfer, i.e., the clothes style of the source person changes to that of the target person. The performance results are shown in Figure \ref{fig8}. In this experiment, we transfer the target person's upper clothes style to the source person by replacing the upper clothes part in $I_{a\_parts}$ of source image. From the results, we can see that our approach can accurately transfer the colors, textures, and styles in clothes while not changing the source person's identity attribute.

\begin{table}
\begin{center}
\setlength{\tabcolsep}{2mm}{
\begin{tabular}{c|c|c|c|c}
\hline \text { Ours } & \text { IS $\uparrow$ } & \text { SSIM $\uparrow$} & \text { FID $\downarrow$} & \text { LPIPS $\downarrow$} \\
\hline 
\hline
\text { w/o MUST} & 3.375 & 0.736 & 21.928 & 0.2840 \\
\text { w/o CA} & \textbf{3.729} & 0.737 & 17.537 & 0.2450 \\
\text { w/o PCM} & 3.649 & 0.724 & 16.963 & 0.2554 \\
\text { Full Model } & 3.692 & \textbf{0.742} & \textbf{15.902} & \textbf{0.2412}\\

\hline
\end{tabular}
}
\end{center}
   \caption{The evaluation results of ablation study.}
\label{tab3}
\end{table}

Further to validate the flexibility of the model, we perform both pose transfer and clothes style transfer at the same time, and the results are shown in Figure \ref{fig1}(c). In the figure, source A's pose is transferred to the target, while its clothes style is replaced by source B's. We can see that the pose transfer and clothes style transfer are both achieved very accurately and realistically. The identity of the generated images remain the same as the source persons, proving that our method can disentangle the appearance and pose well and manipulate them flexibly.

% \begin{figure}
% \begin{center}
% \includegraphics[width=0.8\linewidth]{fig6.pdf}
% \end{center}
%   \caption{The results of achieving pose transfer and clothes style transfer simultaneously.}
% \label{fig6}
% \end{figure}

\subsection{Testing in The Wild}
The person image generation models are trained and tested on the known datasets such as DeepFashion. Still, there are few methods to test the trained models in the wild. In this experiment, we test the performance of our trained model in the wild, and the results are shown in Figure \ref{fig5}. The source images in the figure are obtained from the web. And we use the person parsing method to replace the background of the source images with a clean one so that the complex image background does not affect our approach. As can be seen from the results, our method performs well in maintaining the color and texture of the source person appearance when completing the pose transfer.

\section{Conclusion}
This paper presents a novel multi-level statistics transfer model to realize a self-driven person image generation. The method tackles the hard problem of disentanglement and transfer of appearance and pose attributes in the absence of paired source-target training data. The extraction and transfer of multi-level statistics enable the low-level color and texture information and high-level style information of person images to be well used. Both qualitative and quantitative comparisons demonstrate the superiority of our method. Our approach allows for flexible manipulation of appearance and pose properties to perform pose transfer and clothes style transfer tasks. Finally, our approach also shows the robustness in the wild, which demonstrates the application potential of our model.

\section*{Acknowledgements}
This work was supported by National Natural Science Foundation of China (NSFC) under Grants 61772529, 61972395, 61902400, and Beijing Natural Science Foundation under Grant 4192058.

%-------------------------------------------------------------------------

{\small
\bibliographystyle{ieee_fullname}
\bibliography{egbib}
}

% \appendix

% \section{Additional Implementation Details}
% \label{appendix_A}

\end{document}